\documentclass[10pt,twocolumn,letterpaper]{article}

\usepackage[]{cvpr}      

\usepackage[dvipsnames]{xcolor}

\definecolor{cvprblue}{rgb}{0.21,0.49,0.74}
\usepackage[pagebackref,breaklinks,colorlinks,citecolor=cvprblue]{hyperref}
\usepackage{multirow}
\usepackage{graphicx}
\usepackage{makecell}

\def\paperID{6887} 
\def\confName{CVPR}
\def\confYear{2024}

\title{Unified-Width Adaptive Dynamic Network for All-In-One Image Restoration}

\author{Yimin Xu$^1$, Nanxi Gao$^1$, Yunshan Zhong$^1$, Fei Chao$^1$, Rongrong Ji$^{1*}$\\
$^1$MAC Lab, School of Informatics, Xiamen University \quad  \\
\normalfont{xuyimin9626@gmail.com \quad \{gaonanxi, zhongyunshan\}@stu.xmu.edu.com \{fchao, rrji\}@xmu.edu.cn} \thanks{{$^*$}Corresponding Author}}

\begin{document}
\maketitle
\begin{abstract}
In contrast to traditional image restoration methods, all-in-one image restoration techniques are gaining increased attention for their ability to restore images affected by diverse and unknown corruption types and levels.
However, contemporary all-in-one image restoration methods omit task-wise difficulties and employ the same networks to reconstruct images afflicted by diverse degradations. 
This practice leads to an underestimation of the task correlations and suboptimal allocation of computational resources.
To elucidate task-wise complexities, we introduce a novel concept positing that intricate image degradation can be represented in terms of elementary degradation.
Building upon this foundation, we propose an innovative approach, termed the Unified-Width Adaptive Dynamic Network (U-WADN), consisting of two pivotal components: a Width Adaptive Backbone (WAB) and a Width Selector (WS).
The WAB incorporates several nested sub-networks with varying widths, which facilitates the selection of the most apt computations tailored to each task, thereby striking a balance between accuracy and computational efficiency during runtime.
For different inputs, the WS automatically selects the most appropriate sub-network width, taking into account both task-specific and sample-specific complexities. 
Extensive experiments across a variety of image restoration tasks demonstrate that the proposed U-WADN achieves better performance while simultaneously reducing up to 32.3\% of FLOPs and providing approximately 15.7\% real-time acceleration.
The code has been made available at \url{https://github.com/xuyimin0926/U-WADN}.

\end{abstract}    
\section{Introduction}
\label{sec:intro}

Image restoration is a fundamental endeavor focused on ameliorating the quality of images afflicted by an array of degradations, spanning from noise~\cite{dabov2007color,liu2020lira} to environmental factors such as rain~\cite{fu2017removing,jiang2020multi} and atmospheric haziness~\cite{cai2016dehazenet,qu2019enhanced}.
The importance of image restoration tasks is underscored by their pivotal role in facilitating subsequent downstream applications, notably encompassing image classification~\cite{he2016deep,yu2018slimmable}, object detection~\cite{ren2015faster}, and semantic image segmentation~\cite{ronneberger2015u}.

As deep learning methods have exhibited remarkable efficacy in image restoration tasks, all-in-one image restoration methodologies~\cite{li2022all, park2023all} have gained prominence for their capacity to address multiple degradations in a single model. 
Among these all-in-one methods, corruption-agnostic approaches like AirNet~\cite{li2022all} and IDR~\cite{zhang2023ingredient} are gaining increased attention to relieve the need for predefined degradation types. These methods leverage a degradation encoder to autonomously identify and process the corruption features present in images, thereby circumventing the requirement for any prior assumptions about the nature of the corruption.
Despite achieving commendable performance, these methods process all input samples, characterized by various degradations, within a uniform network structure. 
This seemingly straightforward reconstruction strategy overlooks the task-wise complexities among these diverse restoration tasks, potentially leading to the inefficient allocation of computational resources.
To better characterize task-wise complexity, we posit that one image restoration task can be effectively transformed into another by the deliberate inclusion or removal of corresponding degradation factors. 
For instance, rainy images may be perceived as the introduction of an additional layer of rain atop noisy images, while hazy images can be generated when raindrops are introduced at each pixel of the rainy images, influencing the overall brightness of the original images.
Drawing from this conceptual framework, we can establish a hierarchical ranking of the relative difficulties inherent in each image restoration task; for instance, Deraining $\geq$ Denoising. This ranking allows us to allocate computational resources tailored to each task.
%
%

Building upon the aforementioned analysis, we propose a Unified-Width Adaptive Dynamic Network (U-WADN) method as a means to allocate adaptive computational resources on a task-specific basis.
The U-WADN comprises two core components: the Width-Adaptive Backbone (WAB) and the Width Selector (WS).
The WAB is designed by integrating multiple sub-networks, characterized by a nested structure where smaller sub-networks are encompassed within larger ones. 
%
%
To be more specific, the parameters in a small network are fully reused when generating the reconstructed images from a larger network.
This arrangement mirrors the progression from simpler to more intricate tasks, where intricate image restoration tasks can be transformed by the incorporation of corresponding degradation factors (\emph{i.e.}, extra parameters within larger networks).
%
Upon receiving a pre-trained WAB, the WS plays a pivotal role in selecting the degradation type for incoming input images. It then allocates these images to different sub-networks based on their task-wise complexities.
Furthermore, the WS takes into account sample-wise complexities, where intricate samples within ostensibly straightforward tasks are also processed by larger sub-networks.

The contributions of this paper are summarized as following three-folds:
\begin{itemize}
    \item As far as we know, we are the first work to raise the idea that complicated image restoration tasks can be transformed from simple image restoration tasks. For those decomposable image restoration tasks, we process straightforward image restoration tasks using smaller networks, effectively mitigating the overall computational expenses.
    \item In our study, we introduce a Unified-Width-Adaptive Dynamic Network (U-WADN) designed for all-in-one image restoration tasks. This network comprises a Width-Adaptive Backbone (WAB) and a Width-Selector (WS). The WAB, with its nested network architecture, and the WS, responsible for network selection based on both task-specific and sample-specific complexities, collectively contribute to the reduction of computational overhead, all while maintaining network parameter efficiency.
    \item We conduct experiments on 'Noise-Rain-Hazy' restoration and dynamically allocate computations based on both task-wise and sample-wise difficulties. Despite achieving a 15.7\% real-time acceleration and around 32.3\% FLOPS reduction, U-WADN continues to deliver better performance compared with existing methods.
\end{itemize}

\section{Related work}
\subsection{Image restoration}
Image restoration tasks aim to restore clean images from corrupted ones. 
Conventional image restoration techniques~\cite{rudin1992nonlinear, dong2011image, he2010single} rely on manually crafted priors for the reconstruction of deteriorated images.
With the proliferation of learning-based methods, conventional image restoration techniques have been displaced primarily owing to their exceptional performance.
DnCNN~\cite{zhang2017beyond}, for example, leverages a deep neural network for non-linear mapping in noise reduction.
Chen \emph{et al.}~\cite{chen2022simple} streamline the process by eliminating superfluous activation functions in their baseline network for image restoration.
With the flourish of vision transformers, Liang \emph{et al.}~\cite{liang2021swinir} proposes a transformer-based method to efficiently process images at various scales and resolutions with shifted windows.
However, these image restoration methods lack generalization and struggle to handle multiple degradations simultaneously.

\begin{figure*}[!t]
    \centering
    \includegraphics[width=\linewidth]{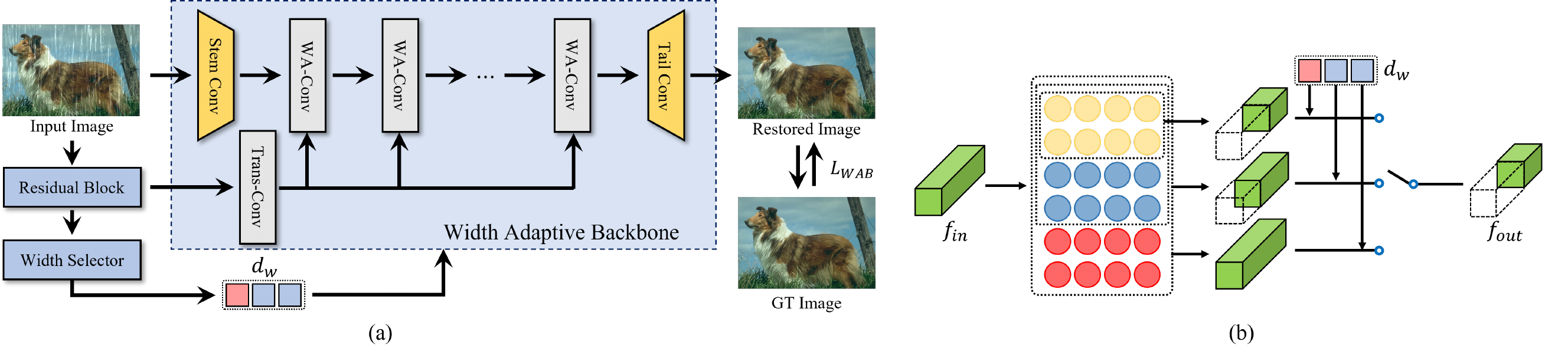}
    \caption{Overall pipeline of the proposed Unified-Width Adaptive Dynamic Network. (a) Unified-Width Adaptive Dynamic Network (U-WADN); (b) Width-Adaptive Convolution.}
    \label{fig:fig2}
\end{figure*}

\subsection{All-in-one image restoration}
Recently, all-in-one image restoration methods are raised to restore multi-degradation inputs within a unified model.
Fan \emph{et al.}~\cite{fan2019general} introduces an innovative weight-adjustable network, allowing for effective reconstruction given the degradation types or corresponding degradation factors. 
Chen \emph{et al.}~\cite{chen2021pre} proposes a transformer-based multi-head and multi-tail structure for different degradation removal. 
However, the above-mentioned methods for all-in-one image restoration necessitate the availability of predefined knowledge to restore corrupted images, such as information regarding the specific degradation type and noise ratio, which is impractical in real-world scenarios.
To mitigate this problem, Li et al.~\cite{li2022all} introduced a Contrast-based degraded Encoder (CE) aimed at capturing distinct features associated with various degradations and leveraging this degradation encoding to facilitate image reconstruction.
Zhang \emph{et al.}~\cite{zhang2023ingredient} further views the various degradation in an ingredient-oriented manner, enhancing the scalability when more image restoration tasks are involved. 
However, these methods use the same network architecture to process corrupted images with various degradations, leading to inefficient and suboptimal use of computational resources.

\subsection{Dynamic neural network in image restoration}
In the realm of image restoration, the concept of dynamic networks has garnered significant attention due to its potential for adaptive resource allocation and enhanced performance. 
ClassSR~\cite{kong2021classsr} introduces a methodology where individual pixels are assigned to distinct convolution branches contingent on their complexities.
Building upon this foundation, Chen \emph{et al.}~\cite{chen2022arm} presented a streamlined model by amalgamating these convolution branches into a versatile, reusable architecture, drawing inspiration from the principles of ClassSR.
Other methodologies, including path selection~\cite{yu2021path} and dynamic selection~\cite{he2020interactive}, have also demonstrated superior efficacy-efficiency tradeoffs in the context of image restoration tasks.
\textcolor{black}{
While existing methods excel in restoring images with single degradation types, our Unified-Width Adaptive Network (U-WADN) is tailored to address more complex all-in-one image restoration challenges involving multiple degradations. 
U-WADN goes beyond assessing just sample-wise difficulties; it also evaluates the intricacies of each restoration task, enabling more effective computational resource allocation.}
\section{Reformulation of Image Degradation}
\label{section:3}
To assess the intricacies of distinct image restoration tasks, we posit that a more complex restoration task can be conceptually formulated as a combination of a simpler restoration task and a residual component. In this section, we delve into the examination of the 'noisy-rain-hazy' scenario as a representative case to elucidate the varying levels of task-wise difficulties encountered in image restoration processes. 
Given that image restoration essentially involves the inverse process of modeling image degradation, we can evaluate the complexities of image restoration tasks by assessing the intricacies encountered in formulating the respective image degradations.
As depicted in previous work~\cite{zhang2023ingredient}, the image degradation process is defined as
\begin{align}
    y = \phi(x;A) + N,
\end{align}
where we denote the corrupted image as $y \in \mathbf{R}^{H\times W\times C}$, while $x \in \mathbf{R}^{H\times W\times C}$ represents the underlying clean image.
$H, W, C$ is the height, width, and number of channels of the images, respectively.
To be more specific,  the corrupted images are generated through the application of a degradation function denoted as $\phi(\cdot, A)$, in conjunction with the addition of noise represented as $N$. 
Here, $A$ signifies the associated degradation parameter.
The noisy images are generated with a straight-forward function $\phi$
\begin{align}
    y_{noise} = x + N,
\end{align}
$y_{noise}$ represents the resultant noisy image. Likewise, in the context of rainy images, the degradation transition $\phi$ is realized through a pixel-wise addition between the clean image $x$ and a rainy image layer denoted as $A_{rain}\in \mathbf{R}^{H\times W\times C}$, formulated as:
\begin{align}
    y_{rain} &= x + A_{rain} + N \nonumber\\
             &= y_{noise} + A_{rain}.
    \label{eq:noise2rain}
\end{align}
When encountering a comparable level of additive noise, it is plausible to perceive the rainy image as the result of combining the noisy image with the rainy image layer.
The hazy image denoted as $y_{haze}$ originates from an element-wise multiplication between the clean image and a hazy degradation map represented as $A_{haze} \in \mathbf{R}^{H\times W\times C}$ as
\begin{align}
    y_{haze} &= x \cdot A_{haze} + N \nonumber\\
             &= x + (A_{haze}-1)\cdot x + N \nonumber\\
             &= y_{noise} + A',
    \label{eq:noise2hazy}
\end{align}
which demonstrates a hazy image can be obtained by adding an instance-specific $A' = (A_{haze}-1)\cdot x$ to the corresponding noisy image. 
In typical scenarios, raindrops are sparse and do not manifest at every pixel in a rainy image, while haze affects all pixels uniformly.
Consequently, the rainy degradation matrix $A_{rain}$ exhibits much sparser characteristics when compared to the hazy degradation matrix $A_{haze}$ and $A'$.
Building upon the aforementioned assumption, the relationship between the hazy image and the rainy image can be further elucidated by extending Equation,\ref{eq:noise2hazy} as:
\begin{align}
    y_{haze} &= y_{noise} + A' \nonumber \\
             &= y_{noise} + A_{rain} + (A' - A_{rain}) \nonumber\\
             &= y_{rain} + (A' - A_{rain})|_{A_{rain}\neq 0} + A'|_{A_{rain} = 0}.
    \label{eq:rain2hazy}
\end{align}
Even when $A'$ is close to $A_{rain}$ at pixels that contain raindrops, \emph{i.e.} $A'|_{A_{rain}\neq 0}$, there is still a need to learn the complementary degradation map $A'|_{A_{rain}=0}$. Hence, the rainy images can be viewed as intermediate results to generate hazy images.
As image restoration is the inverse process of formulating corresponding degradations, we can hierarchically rank task-wise difficulties in the 'noisy-rain-hazy' scenario. For example, restoration of a rainy image is considerably more challenging than that of a noisy image, as it necessitates the incorporation of a residual component, denoted as $A_{rain}$, to effectively restore the rainy image.
Hence, it is convincing to encourage neural networks to share similar behavior and assign more computational resources to more complicated tasks.
\textcolor{black}{It is worth noting that other image restoration tasks, such as image enhancement and image deshadowing, can likewise be conceived as amalgamations of a basic task with a residual component. However, these aspects fall outside the scope of our current paper and are not discussed herein.} 

\begin{figure}
    \centering
    \includegraphics[width=\linewidth]{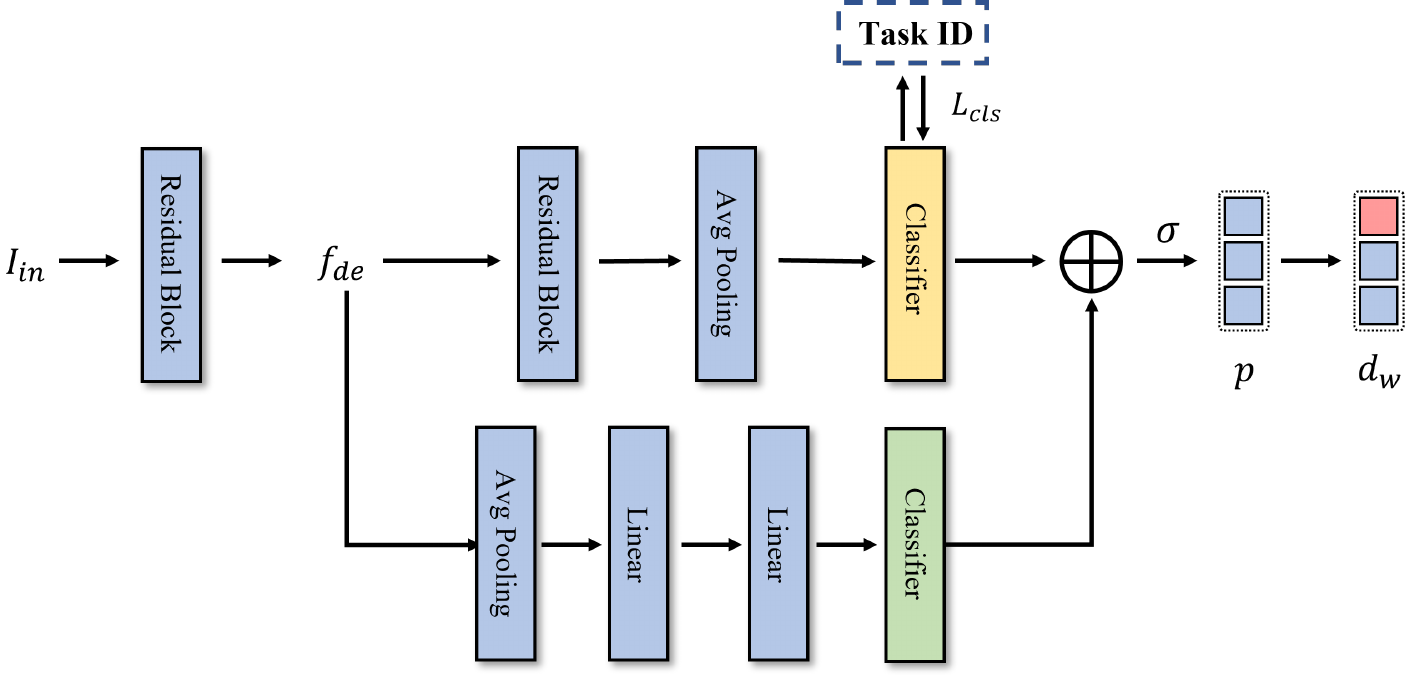}
    \caption{The Width Selector (WS) in the proposed U-WADN.}
    \label{fig:fig3}
\end{figure}

\section{Methods}
In Figure\,\ref{fig:fig2}, we illustrate the architecture of the proposed Unified-Width Adaptive Dynamic Network (U-WADN), which comprises two key components: a Width Adaptive Backbone (WAB) and a Width Selector (WS).
When dealing with corrupted images, the WS takes on the responsibility of dynamically determining the appropriate network width for each individual sample based on both task-specific and sample-specific complexities. Subsequently, the WAB is engaged to perform the restoration of the latent clean images, adjusting its width according to the selections made by WS for each sample.
Given the inherent adaptability of the WAB to various network structures, we have adopted a backbone structure from~\cite{li2022all} in our paper.
To enhance the training efficacy of the U-WADN, we employ a two-stage training strategy. This begins with the initial training of the WAB, followed by the subsequent training of the WS.
In the subsequent section, we will provide a comprehensive exposition elucidating both the WAB and the WS components.

\subsection{Width Adaptive Backbone}

The Width Adaptive Backbone (WAB), is formed by substituting the standard convolutions in~\cite{li2022all} with width adaptive convolutions.
For a clearer illustration of the WAB's functionality, we will concentrate exclusively on the width adaptive convolutions, setting aside other components in our discussion.
Assuming the maximum width, \emph{i.e.} the number of channels, of each convolution in the WAB is denoted as $\omega$, and the weights for each width adaptive convolution are represented as $W \in \mathbf{R}^{\omega \times \omega \times k \times k}$, where $k$ signifies the convolution operation's kernel size and $\omega$ denotes the number of input channels and output channels. To simplify the notation, we assume the number of input channels equals that of output channels.
The sub-networks within the WAB are derived by considering only the initial $\rho$ input and output channels of each width adaptive convolution, where $\rho \in \left[\omega_1, \omega_2, \dots, \omega_n\right]\subseteq (0,\omega]$ represents one of the $n$ width options.
The weight of the width adaptive convolution within the sub-network can be expressed as $W' \in \mathbf{R}^{\rho \times \rho \times k \times k}$.
Following the successful training of the WAB, we obtain a total of $n$ sub-networks, with the width ranging from $\omega_1$ to $\omega_n$.

\textcolor{black}{In the context of all-in-one image restoration methods, the necessity to encapsulate the degradation features, denoted as $f_{de} \in \mathbf{R}^{C \times H \times W}$, from individual samples mandates the utilization of a residual block. To align the dimensional integrity of the degradation encoding with that of the width adaptive convolutions, we implement a transformation convolution characterized by a weight  $W_{trans} \in \mathbf{R}^{C \times \rho \times 1 \times 1}$ to obtain $f'_{de}\in \mathbf{R}^{\rho \times H \times W}$.}

\textbf{Correlation between the WAB and restoration tasks reformulation.} As delineated in Section\,\ref{section:3}, the process of restoring a complicated degradation entails the restoration of a fundamental degradation component in conjunction with its associated residuals.
Within this section, we endeavor to illustrate that the concept of the Width Adaptive Backbone aligns with this underlying assumption.
For a small-width network with width $\omega_1$, suppose the input feature map $f_{i} \in \mathbf{R}^{\omega_1\times H \times W}$ in the layer $i$, the output feature map of this layer $f_{i+1} \in \mathbf{R}^{\omega_1\times H \times W}$ can be denoted as
\begin{align}
    f_{i+1} = W_{i}[0:\omega_1,0:\omega_1, :, :]*f_{i},
\end{align}
where $W_i\in \mathbf{R}^{\omega \times \omega \times k\times k}$ is the weight of layer $i$ and $*$ denotes the convolution operation.
As the width operator only works in the first two dimensions of the weight matrix $W_{i}$, we simplify the convolution operation by
\begin{align}
    F_{i+1} = \hat{W_i}[0:\omega_1,0:\omega_1] \times F_{i},
\end{align}
where $F_{i}$ is the convolution results between $f_{i}$ and corresponding convolution kernels in $W_i$. 
$\hat{W_i}$ is the simplified weight matrix in the layer $i$.
\begin{figure*}[!t]
    \centering
    \includegraphics[width=\linewidth]{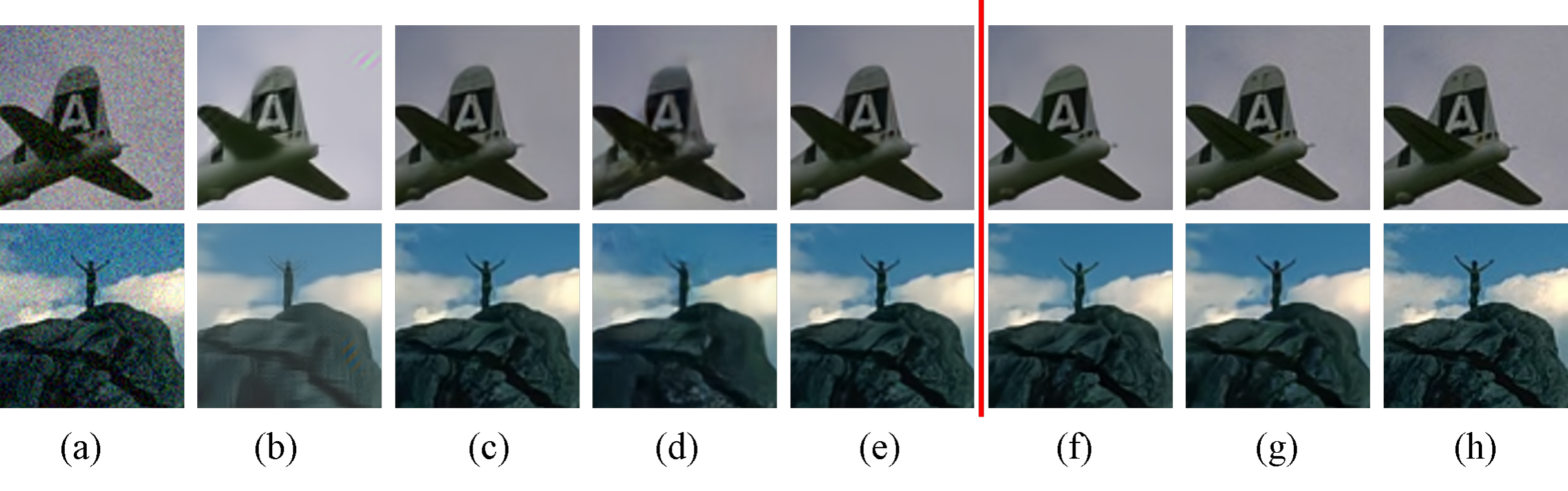}
    \caption{Visual comparison among methods on image denoising. (a) is the input noisy image, and from (b) to (g) are results from  CBM3D~\cite{dabov2007color}, DnCNN~\cite{zhang2017beyond}, IRCNN~\cite{zhang2017learning}, FFDNet~\cite{zhang2018ffdnet}, AirNet~\cite{li2022all} and U-WADN, respectively. (h) is the ground truth of clean images.
}
    \label{fig:denoise}
\end{figure*}
Hence, for the network with a larger width $\omega_2$, the output feature map can be denoted as
\begin{align}
    F'_{i+1} & = \hat{W_i}[0:\omega_2, 0:\omega_2] \times F'_i  \nonumber \\
    &= \begin{bmatrix} F_{i+1}\\ 0\end{bmatrix} + O_{i},
    \label{equation:sub}
\end{align}
where $O_i$ denotes the reminder at layer $i$. We can then move one step forward, the output feature map in layer $i+2$ can be denoted as
\begin{align}
    F'_{i+2} = \hat{W_{i+1}}[0:\omega_2,0:\omega_2] \times F'_{i+1}.
    \label{equation:ori}
\end{align}
Substitute $F'_{i+1}$ in Eq\,\ref{equation:ori} with Eq\,\ref{equation:sub}, we can get
\begin{align}
    F'_{i+2} = \begin{bmatrix} F_{i+2}\\ 0\end{bmatrix} + O_{i+1}.
    \label{equation:conclusion}
\end{align}
Equation\,\ref{equation:conclusion} demonstrates that the output features of a larger sub-network $F'$ can always be decomposed to the combination of the output features of the small sub-network and a remainder $O$.
Therefore, the process of restoring more complexly corrupted images using larger sub-networks within the WAB can be equated to the restoration of simpler corrupted images with smaller sub-networks, supplemented by the remainder. This approach aligns with the assumptions outlined in Section\,\ref{section:3}. Please refer to the supplementary material for detailed proofs. 

\textbf{The training objective of the WAB.}  For each iteration, we randomly select a sub-network with the width $\rho \in [\omega_1, \omega_2, \dots, \omega_{n-1}]$, and the full network with the width $\omega_n$ to balance the training of each sub-networks.
$\mathcal{L}_1$ loss is calculated between the restored images and ground truth clean images for selected sub-networks.
To be more specific, the loss can be represented by
\begin{align}
    L_{recon} = \mathcal{L}_1(S(I;\theta_{:\rho}), Y) + \mathcal{L}_1(S(I;\theta), Y),
\end{align}
where $S(\cdot, \theta)$ is the sub-network parameterized by $\theta$ and $Y$ is the ground truth clean image. To further enhance the consistency of the output from different sub-networks, we introduce a distillation loss as,
\begin{align}
     L_{distill} = \mathcal{L}_1(S(I;\theta_{:\rho}), S(I;\theta)).
\end{align}
To better capture the degradation encoding $f_{de}$, it is passed to a classification network $N$ with parameter $\zeta$ and supervised by a classification loss
\begin{align}
    L_{de} = CE(N(f_{de};\zeta), y),
\end{align}
where  $y$ is the type of task and $CE$ denotes the cross entropy loss. 
Hence, the training object of the proposed WAB is defined as
\begin{align}
    L_{WAB} = L_{recon} + L_{distill} + L_{de}.
    \label{eq:distill}
\end{align}

\begin{figure*}[th]
    \centering
    \includegraphics[width=\linewidth]{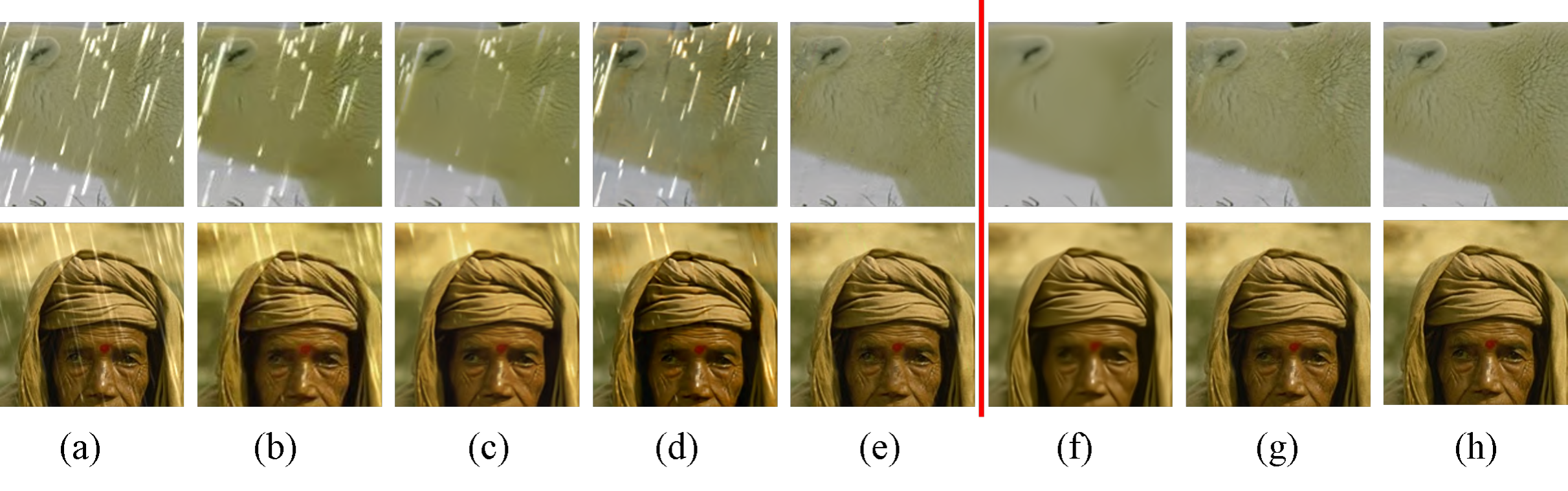}
    \caption{Visual comparison among methods on image deraining. (a) is the input rainy image, and from (b) to (g) are results from  DIDMDN~\cite{zhang2018density}, UMRL~\cite{yasarla2019uncertainty}, SIRR~\cite{wei2019semi}, LPNet~\cite{fu2019lightweight}, AirNet~\cite{li2022all}, and U-WADN, respectively. (h) is the ground truth of clean images.}
    \label{fig:rain}
\end{figure*}

\begin{table*}[th]
\resizebox{\linewidth}{!}{
\begin{tabular}{c|ccc|c|c|c}
\hline
\multirow{2}{*}{Method} & \multicolumn{3}{c|}{Denoise} & Derain & Dehaze & \multirow{2}{*}{Average} \\ \cline{2-6} 
 & BSD68 ($\sigma=15$) & BSD68 ($\sigma=25$) & BSD68 ($\sigma=50$) & Rain100L & SOTS &  \\ \hline
BRDNet~\cite{tian2020image}  & 32.26/0.8977 & 29.76/0.8355 & 26.34/0.6934 & 27.42/0.8952 &  23.23/0.8952 & 27.80/0.8434 \\
LPNet~\cite{fu2019lightweight}&26.47/0.7780 & 24.77/0.7477 & 21.26/0.5522 & 24.88/0.7837 & 20.84/0,8277 & 23.64/0.7379 \\
FDGAN~\cite{dong2020fd}& 30.25/0.9103 & 28.81/0.8682 & 26.43/0.7757 & 29.89/0.9329 & 24.71/0.9294 & 28.02/0.8833 \\
MPRNet~\cite{zamir2021multi} & 33.54/0.9274 & 30.89/0.8797 & 27.56/0.7792 & 33.57/0.9542 & 25.28/0.9545 & 30.17/0.8990 \\
DL~\cite{fan2019general} & 33.05/0.9140 &  30.41/0.8606 & 26.90/0.7401 & 32.62/0.9314 & 26.92/0.9314 & 29.98/0.8755 \\
AirNet~\cite{li2022all} & \textbf{33.92/0.9329} & \textbf{31.26/0.8884} & \textbf{28.00/0.7974} & \underline{34.90/0.9675} & \underline{27.94/0.9615} & \underline{31.20/0.9095}\\
U-WADN & \underline{33.73/0.9311} & \underline{31.14/0.8864} & \underline{27.92/0.7932} & \textbf{35.36/0.9684} & \textbf{29.21/0.9709} &  \textbf{31.47/0.9097}\\ \hline
\end{tabular}
}
\caption{Quantitative results on five image restoration tasks with all-in-one methods. The best and second best results are marked in \textbf{bold} and \underline{underline}, respectively.}
\vspace{-1em}
\label{table:1}
\centering
\end{table*}

\subsection{Width Selector}
After obtaining a pre-trained Width Adaptive Backbone, the Width Selector (WS) takes on the responsibility of dynamically selecting the appropriate width, taking into account both task-specific and sample-specific complexities.
For an input image $I_{in} \in \mathbf{R}^{3\times H\times W}$, a residual block first captures its degradation encoding $f_{de} \in \mathbf{R}^{C\times H\times W}$.
As illustrated in Figure\,\ref{fig:fig3}, the Width Selector consists of two branches. The upper branch is tasked with extracting task-specific difficulties, while the lower branch focuses on capturing sample-specific input features.

As for the task-specific branch, the $f_{de}$ goes through a residual block and an average pooling layer to capture the task-specific logits $f_{task}$ as
\begin{align}
    f_{task} = AvgPool(R(f_{de}; \phi_1))),
\end{align}
where $R(\cdot; \phi)$ is the residual blocks parameterized with $\phi$. Similarly, the sample-specific logits are obtained by 
\begin{align}
    f_{sample} = L(L(AvgPool(f_{de});\gamma_1);\gamma_2).
\end{align}
$L(\cdot;\gamma)$ is the linear projection whose weight is $\gamma$.  Then the task-specific and sample-specific logits go through corresponding classifiers with parameter $\xi$, \emph{i.e.} $C(\cdot, \xi_1)$ and $C(\cdot, \xi_2)$, to obtain the decision logits
\begin{align}
    f_{decision} = C(f_{task}; \xi_1) + C(f_{sample}; \xi_2).
\end{align}
The possibility of choosing each width candidate, \emph{i.e.} $p_i \in [p_0, p_1, \dots, p_n]$, can be obtained by passing $f_{decision}$ through a softmax activation function. The discrete decision $d_w$ is the width candidate with the largest $p_i$.

\textbf{Training Objective for WS.} The training loss for WS consists of a classification loss $L_{cls}$, a sparsity loss $L_{spars}$, and a selection loss $L_{select}$
\begin{align}
    L_{WS} = L_{cls} + L_{spars} + L_{select}.
\end{align}
The classification loss strengthens the correlation between the task-specific logits and the corresponding task as
\begin{align}
    L_{cls} = CE(C(f_{task};\xi_{1}), y),
\end{align}
The sparsity loss forces the WS to select a smaller network instead of processing all samples with the largest backbone
\begin{align}
    L_{spars} = \frac{1}{m}\sum_{j=1}^{mn}\left(\sum_{i=0}^{n}\frac{p_i^j\omega_{i}}{\omega} - t\right)^2.
    \label{eq:target}
\end{align}
$t \in [0,1]$ is the target width ratio and $m$ is the number of samples per batch. The selection loss aims to minimize the reconstruction loss after width selection. To be more specific, the $\mathcal{L}_1$ loss of each sub-network can be derived as $[L_1,\dots, L_n]$. Hence, the selection loss is represented as
\begin{align}
    L_{select} = \frac{1}{mn}\sum_{j=1}^{m}\sum_{i=0}^{n}p_i^jL_i^j.
\end{align}
\begin{figure*}[th]
    \centering
    \includegraphics[width=\linewidth]{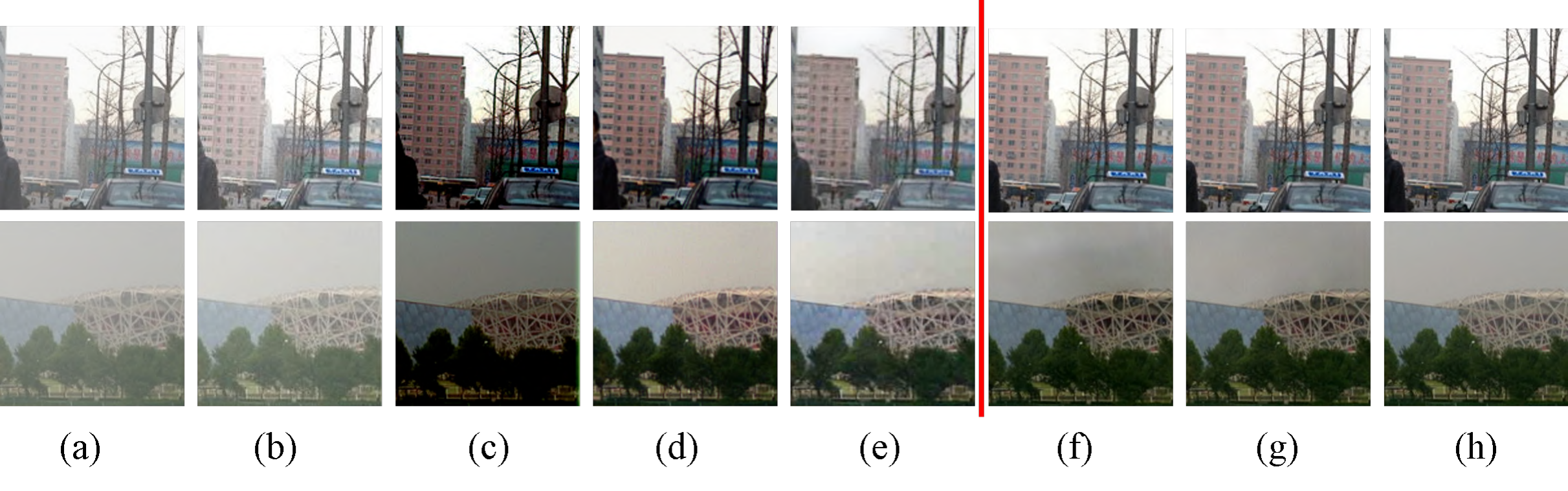}
    \caption{Visual comparison among methods on image dehazing. (a) is the input rainy image, and from (b) to (g) are results from  DehazeNet~\cite{cai2016dehazenet}, AOD-Net~\cite{li2017aod}, EPDN~\cite{qu2019enhanced}, FDGAN~\cite{dong2020fd}, AirNet~\cite{li2022all}, and U-WADN, respectively. (h) is the ground truth of clean images.}
    \label{fig:hazy}
    \vspace{-1em}
\end{figure*}

\begin{figure*}[!t]
    \begin{minipage}{0.48\textwidth}
        \centering
        \includegraphics[width=\linewidth]{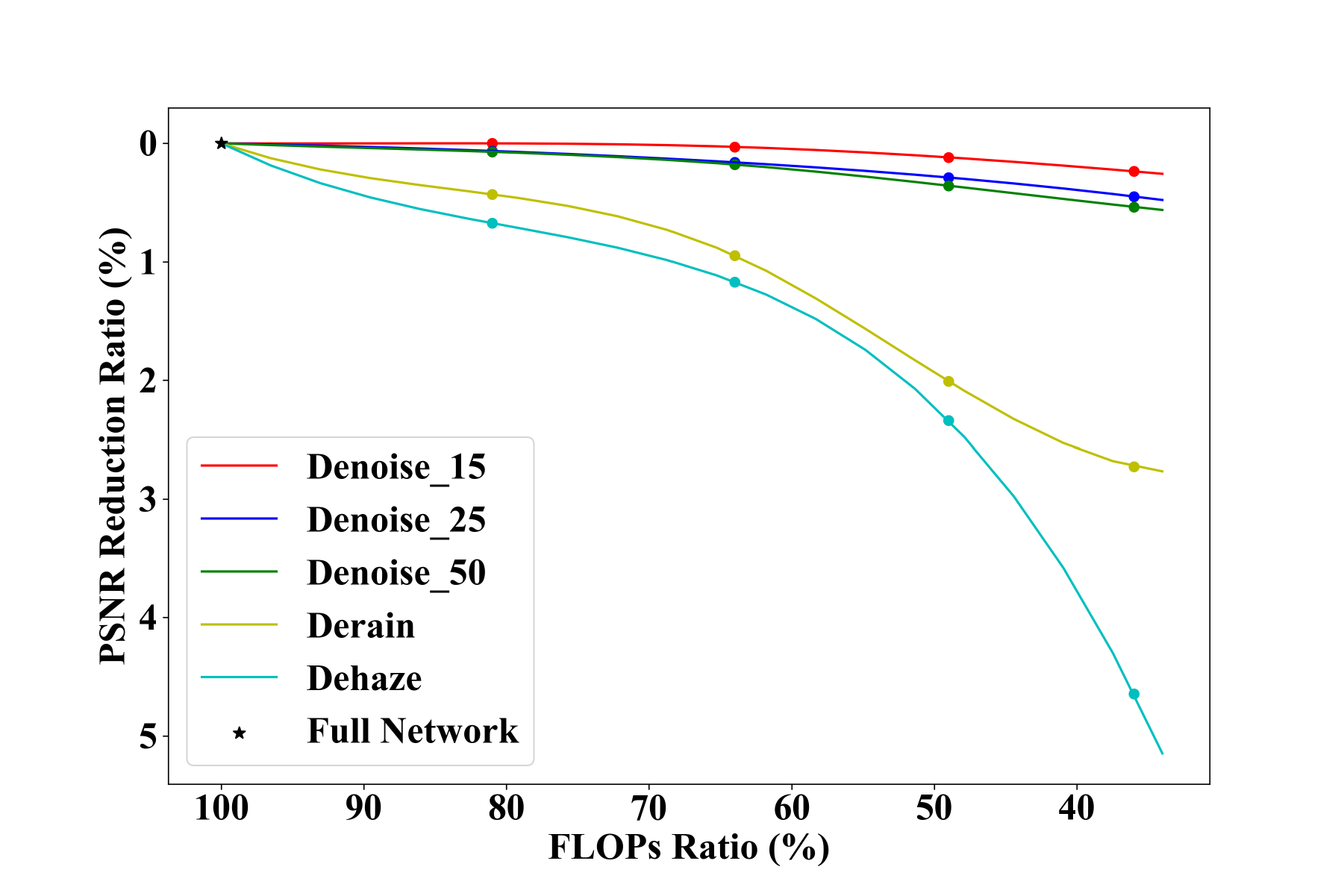}
        \caption{The relationship between the PSNR Reduction Ratio and FLOPs Ratio of various image degradation tasks.}\label{fig:1}
   \end{minipage}\hfill
   \begin{minipage}{0.48\textwidth}
        \centering
        \includegraphics[width=\linewidth]{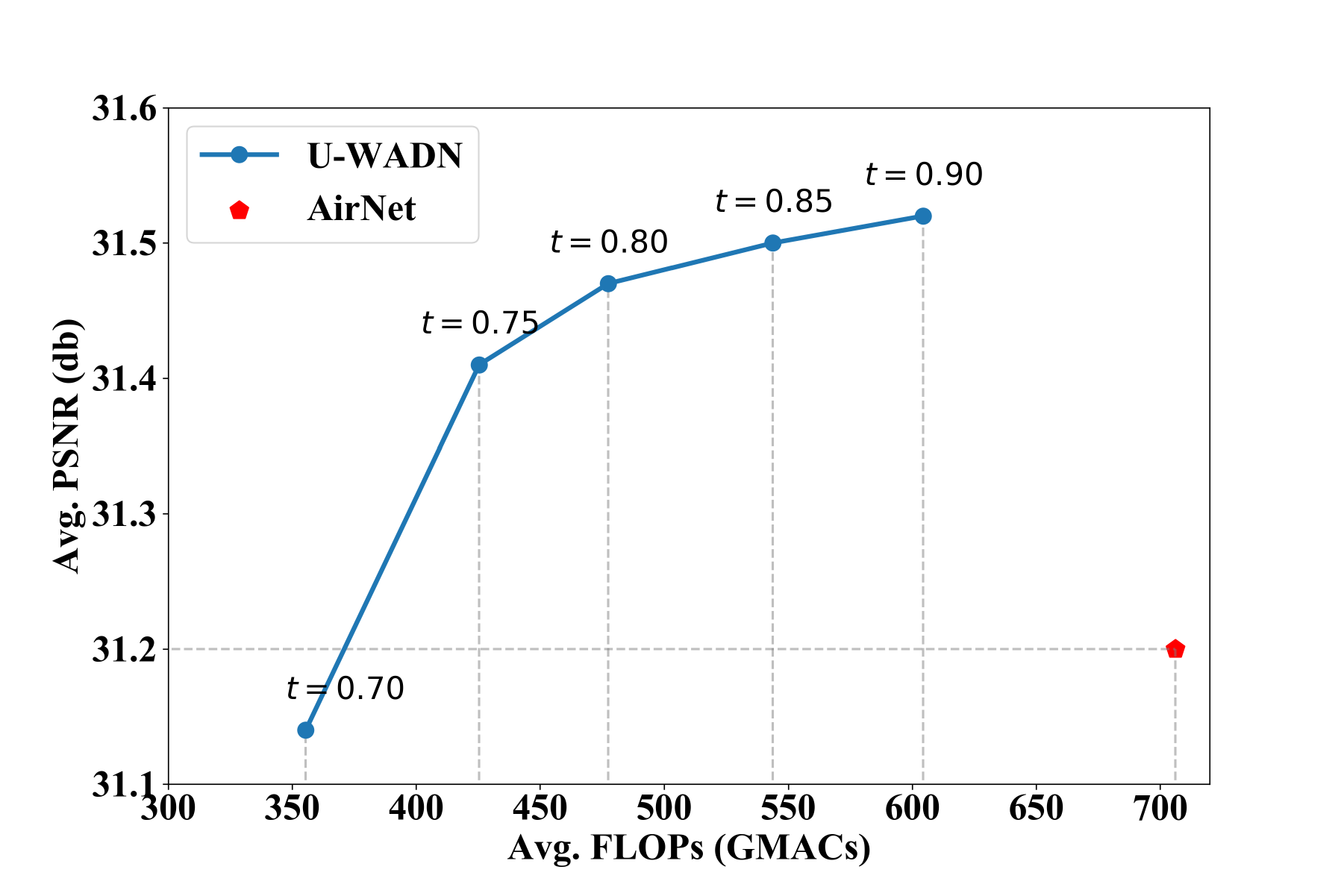}
         \caption{Average PSNR and average FLOPs with respect to sparsity target $t$.}\label{fig:8}
   \end{minipage}
   \vspace{-1em}
\end{figure*}

\begin{table*}[ht]
\centering
\resizebox{\linewidth}{!}{
\begin{tabular}{c|ccccc|c|c|c|c}
\hline
\multirow{2}{*}{Method} & \multicolumn{5}{c|}{Avg. Width Ratio $(\rho/\omega)$} & \multirow{2}{*}{Avg. PSNR} & \multirow{2}{*}{Parameters (M)} & \multirow{2}{*}{FLOPs (G)} & \multirow{2}{*}{Avg. Time (s)} \\ \cline{2-6} 
& \makecell[c]{Denoise \\ $\sigma=15$} & \makecell[c]{Denoise \\ $\sigma=25$} & \makecell[c]{Denoise \\ $\sigma=50$} & Derain & Dehaze&  &  &  &  \\ \hline
AirNet (Baseline) & 1.00 & 1.00 & 1.00 & 1.00 & 1.00 & 31.20 & 5.77 & 706.11 & 0.32 \\ \hline
\multirow{5}{*}{\begin{tabular}[c]{@{}c@{}}U-WADN\\ (without Width Selector)\end{tabular}} 
 & 0.60 & 0.60 & 0.60 & 0.60 & 0.60 & 30.80 & 5.77 & 257.20 & 0.20 ($\uparrow 37.5\%$)\\
 & 0.70 & 0.70 & 0.70 & 0.70 & 0.70 & 31.08 & 5.77 & 350.89 & 0.23 ($\uparrow 28.1\%$) \\
 & 0.80 & 0.80 & 0.80 & 0.80 & 0.80 & 31.30 & 5.77 & 458.51 & 0.26 ($\uparrow 18.8\%$) \\
 & 0.90 & 0.90 & 0.90 & 0.90 & 0.90 & 31.47 & 5.77 & 580.05 & 0.29 ($\uparrow 9.40\%$)\\ 
 & 1.00 & 1.00 & 1.00 & 1.00 & 1.00 & 31.60 & 5.77 & 706.11 & 0.32 ($\uparrow 0.00\%$) \\\hline
\begin{tabular}[c]{@{}c@{}}U-WADN\end{tabular} & 0.60 & 0.70 & 0.79 & 0.92 & 0.99 & 31.47 & 6.05 & 477.51 & 0.27 ($\uparrow 15.7\%$) \\ \hline
\end{tabular}
}
\caption{Quantitative results on five image restoration tasks with different widths. The width ratio signifies the proportion of the maximum width utilized for a given task. For instance, $\rho/\omega = 0.6$ means only initial $0.6\omega$ channels are adopted in the WAB.}
\vspace{-1em}
\label{table:2}
\end{table*}

\begin{figure*}
    \centering
    \includegraphics[width=1.0\linewidth]{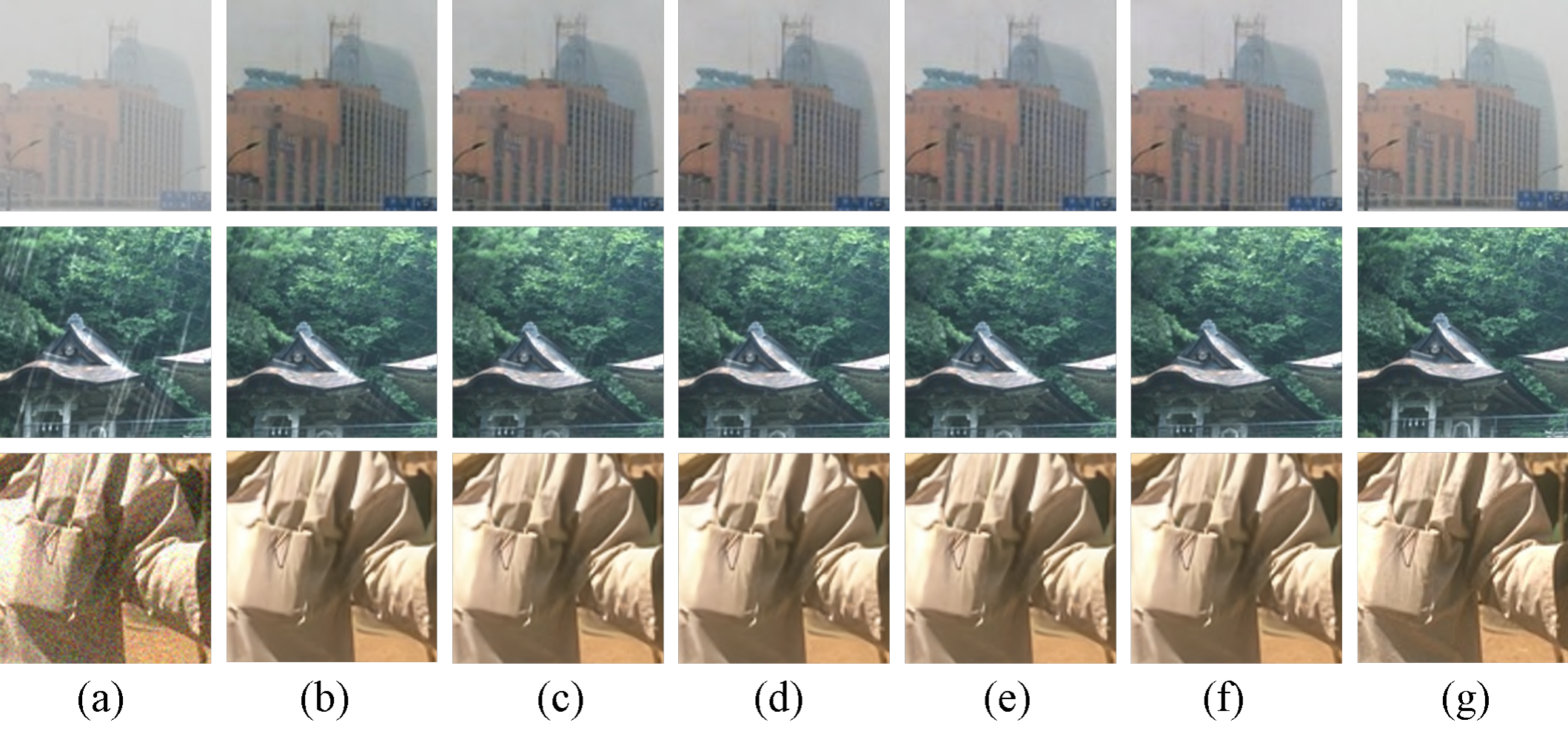}
    \caption{Visual comparison among methods on different sub-networks. (a) is the input noisy image, and from (b) to (f) are results from a width ratio of 0.6-1.0, respectively. (g) is the ground truth of clean images. The first row is the dehazing result, the second row is the deraining result and the last row is the denoising result.}
    \label{fig:0.6-1.0}
\end{figure*}

\section{Experiments}
\subsection{Experimental Settings}
\textbf{Datasets and evaluation metrics.} Building upon the work of~\cite{li2022all}, we undertake comprehensive experimental evaluations across five distinct image restoration tasks, specifically focusing on image denoising at noise levels $\sigma = 15, 25, 50$, image deraining, and image dehazing.
Consequently, our experimentation necessitates the utilization of five corresponding datasets, which comprise BSD400, BSD68~\cite{martin2001database}, WED~\cite{ma2016waterloo} for the purpose of image denoising, Rain100L~\cite{yang2019joint} for image deraining, and RESIDE~\cite{li2018benchmarking} for image dehazing.
To be more specific, we follow~\cite{li2022all} of the division of training and testing sets. For image denoising, BSD400 and WED~\cite{ma2016waterloo} are used for training while BSD68~\cite{martin2001database} is the testing set with 68 ground truth images. For image deraining, we adopt the Rain100L~\cite{yang2019joint} dataset which consists of 200 rainy-clean training pairs and 100 testing image pairs. For image dehazing, we conduct the experiments on the RESIDE~\cite{li2018benchmarking} dataset with the Outdoor Training Set (OTS) and the Synthetic Objective Testing Set (SOTS) for training and testing, respectively.
We employ the Peak Signal-to-Noise Ratio (PSNR) and the Structural Similarity Index Measure(SSIM) to provide a quantitative assessment of the image restoration models' performance.
Besides the performance of each task, we also report the average PSNR/SSIM among all tasks, as the task types are often unattainable in real-world scenarios.
Additionally, we have included the number of parameters, Floating-Point Operations (FLOPs), and execution times as metrics to gauge the efficiency of the proposed methods.

\textbf{Training Details.} The Unified-Width Adaptive Dynamic Network (U-WADN) is trained on 4 NVIDIA GeForce RTX 3090 GPUs.
As the U-WADN is adaptive to all network structures, we simply borrow the network from AirNet~\cite{li2022all} to demonstrate the efficacy and efficiency of the proposed methods.
For the training of Width-Adaptive Backbone (WAB), the width candidates $\rho$ of WAB are set to be $[0.6\omega, 0.7\omega, 0.8\omega, 0.9\omega, 1.0\omega]$ to meet the requirement of 5 image restoration tasks, where the maximum width $\omega$ is set to 64. 
The WAB is trained for 2000 epochs with the Adam optimizer. 
After the WAB is trained properly, the Width Selector (WS) is trained for another 20 epochs with a learning rate of 0.01.

\begin{table}[t]
\centering
\resizebox{\linewidth}{!}{
\begin{tabular}{ccc|c}
\hline
\multicolumn{3}{c|}{Method} & \multirow{2}{*}{PSNR/SSIM} \\ \cline{1-3}
\multicolumn{1}{c|}{Plain} & \multicolumn{1}{c|}{w/ distillation} & w/ transform &  \\ \hline
\multicolumn{1}{c|}{\checkmark} & \multicolumn{1}{c|}{} &  & 30.98/0.9057\\ 
\multicolumn{1}{c|}{\checkmark} & \multicolumn{1}{c|}{\checkmark} &  & 31.35/0.9092 \\
\multicolumn{1}{c|}{\checkmark} & \multicolumn{1}{c|}{} & \checkmark & 31.31/0.9096 \\ 
\multicolumn{1}{c|}{\checkmark} & \multicolumn{1}{c|}{\checkmark} & \checkmark & 31.47/0.9097 \\ \hline
\end{tabular}
}
\caption{Ablation experiments on the designed WAB components.}
\vspace{-1.5em}
\label{table:3}
\end{table}
\subsection{Comparison with state-of-the-art methods}

Building upon the foundations laid by Li et al.\cite{li2022all}, our analysis positions the novel U-WADN framework with six state-of-the-art image restoration methodologies, namely, BRDNet\cite{tian2020image}, LPNet~\cite{fu2019lightweight}, FDGAN~\cite{dong2020fd}, MPRNet~\cite{zamir2021multi}, DL~\cite{fan2019general}, and AirNet~\cite{li2022all}.
To ensure a fair comparison, we re-implement the first five methods in an all-in-one manner.
As consolidated in Table,\ref{table:1}, the U-WADN achieves a superior mean PSNR/SSIM metric by \textbf{0.27} relative to the current state-of-the-art (SOTA) AirNet benchmark. 
The benefits conferred by the U-WADN are particularly salient in the restoration of complex tasks such as image deraining and dehazing, enhancing the PSNR by \textbf{0.46} and \textbf{1.27} respectively.
The visual comparison, as depicted in Figures\,\ref{fig:denoise}-\,\ref{fig:hazy}, further substantiates the perceptually superior performance of the U-WADN. 
When contrasted with both singular degradation restoration methods and all-in-one image restoration techniques — demarcated by a red delineation — the imagery reconstructed by the U-WADN is richer in contextual details and exhibits more defined edges.

In addition to its superior performance capabilities, the U-WADN demonstrates significant efficiency improvements, achieving a reduction in average FLOPs by \textbf{32.3\%} and a \textbf{15.7\%} acceleration in processing speed according to Table\,\ref{table:2} compared with the AirNet baseline.
Furthermore, the U-WADN achieves performance comparable to the full network with just 60\% of the maximum width for image denoising tasks and surpasses the AirNet baseline in image deraining and dehazing without utilizing the full network.

\subsection{Ablation Studies}

\textbf{Effectiveness of the WS.} As indicated in Table\,\ref{table:2}, in the absence of the Width Selector (WS), the U-WADN is constrained to applying an identical width to all processed corrupted input samples
However, the integration of the WS into the U-WADN architecture significantly enhances the model's ability to balance efficacy and efficiency, effectively addressing both the nuances of different tasks and the unique characteristics of each sample.
When operating at comparable levels of FLOPs, the U-WADN equipped with the WS demonstrates superior performance over its counterpart without the WS, \emph{i.e.} U-WADN with an identical width of 0.8, achieving an average PSNR increase of \textbf{0.17}. Moreover, in comparison to the U-WADN set to an identical width of 0.9, the implementation of the WS yields a substantial \textbf{20\%} reduction in computational requirements without sacrificing performance. 
Notably, the U-WADN equipped with the WS primarily concentrates on task-wise challenges while also considering sample-wise difficulties, especially in tasks like image deraining and dehazing.

\textbf{Efficacy of components in the WAB.} The efficacious training of the Width-Adaptive Backbone (WAB) hinges on two pivotal elements: the distillation loss delineated in Equation,\ref{eq:distill} and the transformation block illustrated in Fig,\ref{fig:1}. 
To maintain integrity and uniformity across experimental conditions, the training regimen deployed for the Width Selector (WS) is uniformly applied across all ablation studies.
Empirical data presented in Table,\ref{table:3} corroborate that the incorporation of the distillation loss and the transformation block substantively augments the WAB's capability. 
The distillation loss impels the smaller sub-networks to emulate the reconstruction patterns of their more extensive counterparts, while the transformation block is instrumental in the precise capture of degradation encodings within the sub-networks.
The amalgamation of the distillation loss with the transformation block catalyzes the WAB to attain its optimal performance parameters.

\textbf{Analysis of task-wise difficulty.} As substantiated in Section\,\ref{section:3}, a theoretical analysis of the relative difficulty associated with diverse image restoration tasks has been conducted. In this discourse, we expound upon a pragmatic approach to quantifying the challenges of each task.
As shown in Fig\,\ref{fig:1}, while the task of image denoising retains a comparatively elevated PSNR even when operating at a mere 40\% of full FLOPs, there is a marked and precipitous decline in PSNR observed in the context of image dehazing.
This empirical observation permits a hierarchical classification of task-wise difficulty, where image dehazing is the most challenging image restoration task among these 5 tasks.
Furthermore, the visualization results in Fig\,\ref{fig:enter-label} can also demonstrate the different complexities of each task. To be more specific, there is an obvious visual enhancement in image dehazing when the width of the sub-network increases from 0.6 to 1.0. While the rain-drop removal shows a similar but less obvious trend, the image denoising shows a subtle change from width 0.6 to width 1.0. Based on the visualization results, we can also drive to the conclusion that the relative difficulties of these tasks are image dehazing $\geq$ image deraining $\geq$ image denoising.

\textbf{Influence of sparsity target $t$.}The parameter $t$ as defined in Equation,\ref{eq:target} serves as a regulatory mechanism for the trade-off between efficacy and efficiency.
As shown in Fig\,\ref{fig:8}, an increased sparsity target is associated with enhanced performance at the expense of greater computational overhead. 
In pursuit of an optimal equilibrium between computational expenditure and performance output, our experimental framework adopts a sparsity target fixed at 0.8.

\section{Conclusion}

In this paper, we introduced the Unified-Width Adaptive Network (U-WADN) for the all-in-one image restoration task, which dynamically allocated computational resources by considering both the task-wise and sample-wise difficulties.
Comprehensive experiments demonstrated that the U-WADN not only achieved higher PSNR but also achieved lower FLOPs and enhanced inference speed compared to current all-in-one image restoration benchmarks.
As U-WADN is compatible with a wide range of image restoration approaches and tasks,  the U-WADN can be widely applied across various restoration tasks in future research.
{
    \small
    \bibliographystyle{ieeenat_fullname}
    \bibliography{main}
}


\end{document}